\documentclass{article} %
\usepackage[preprint]{colm2025_conference}

\usepackage{amsmath}
\usepackage{microtype}
\usepackage{hyperref}
\usepackage{cleveref}
\usepackage{url}
\usepackage{booktabs}
\usepackage{graphicx}
\usepackage{subfigure}

\usepackage{colortbl}
\usepackage{multirow}
\usepackage{lineno}
\usepackage{makecell}

\usepackage{algorithm}
\usepackage{algpseudocode}
\usepackage{amsmath}
\usepackage{amssymb}

\usepackage{xcolor}
\definecolor{darkblue}{RGB}{0,0,128}
\definecolor{codeblue}{RGB}{45, 117, 182}
\definecolor{codepurple}{RGB}{109, 53, 193}
\definecolor{codegreen}{RGB}{84, 130, 53}
\definecolor{codered}{RGB}{238, 34, 12}

\definecolor{textgray}{RGB}{145,145,145}
\definecolor{backgray}{RGB}{221,221,221}

\hypersetup{colorlinks=true, citecolor=darkblue, linkcolor=darkblue, urlcolor=darkblue}

\newcommand{\method}{\textsc{Pacer}}

\title{\method{}: Blockwise Pre-verification for Speculative Decoding with Adaptive Length}

\author{Situo Zhang$^1$\thanks{Equal contribution.}, Yifan Zhang$^{1*}$, 
Zichen Zhu$^1$, Hankun Wang$^1$, Da Ma$^1$, \\\textbf{Danyang Zhang}$^1$,
\textbf{Lu Chen}$^{1,2,3}$, \textbf{Kai Yu}$^{1,2,3}$ \\
\textsuperscript{1}X-LANCE Lab, School of Computer Science\\MoE Key Lab of Artificial Intelligence, SJTU AI Institute\\Shanghai Jiao Tong University, Shanghai, China\\
\textsuperscript{2}Jiangsu Key Lab of Language Computing, Suzhou, China\\
\textsuperscript{3}Suzhou Laboratory, Suzhou, China\\
}

\begin{document}

\ifcolmsubmission
\linenumbers
\fi

\maketitle

\begin{abstract}
Speculative decoding (SD) is a powerful technique for accelerating the inference process of large language models (LLMs) without sacrificing accuracy. Typically, SD employs a small draft model to generate a fixed number of draft tokens, which are then verified in parallel by the target model. However, our experiments reveal that the optimal draft length varies significantly across different decoding steps. This variation suggests that using a fixed draft length limits the potential for further improvements in decoding speed. To address this challenge, we propose \method{}, a novel approach that dynamically controls draft length using a lightweight, trainable pre-verification layer. This layer pre-verifies draft tokens blockwise before they are sent to the target model, allowing the draft model to stop token generation if the blockwise pre-verification fails. We implement \method{} on multiple SD model pairs and evaluate its performance across various benchmarks. Our results demonstrate that \method{} achieves up to \(2.66\times\) Speedup over autoregressive decoding and consistently outperforms standard speculative decoding. Furthermore, when integrated with Ouroboros, \method{} attains up to \(3.09\times\) Speedup.
\end{abstract}

\section{Introduction}
Large language models (LLMs) have revolutionized artificial intelligence in recent years~\cite{achiam2023gpt, guo2025deepseek, yang2025qwen3}, demonstrating exceptional capabilities across a wide range of tasks, including conversational agents~\cite{chiang2024chatbot}, code generation~\cite{jimenez2024swebench}, and complex reasoning~\cite{luong2025towards}. Despite their impressive performance, the autoregressive nature of current LLMs, which generate tokens sequentially, introduces substantial inference latency. This latency is primarily caused by memory bandwidth constraints (\textbf{memory bound}) rather than computational limitations (\textbf{compute bound})~\cite{shazeer2019fast, cai_medusa_2024}, leading to underutilization of GPU parallelism. As a result, LLMs remain difficult to deploy in time-critical applications and on resource-constrained edge devices~\cite{Stern_Blockwise_2018, ivanov2021data}.

Numerous methods have been proposed to address this challenge, among which speculative decoding~(SD) stands out as an effective technique for accelerating LLM inference~\cite{Stern_Blockwise_2018, leviathan_fast_2023, chen_accelerating_2023}. The core idea involves splitting the decoding process into two stages: drafting and verification. A small and efficient draft model predicts a sequence of \(\gamma\) tokens in advance, which are then verified in a single forward pass by the larger target model. This approach preserves the quality of the generated output while improving inference speed.

\begin{figure}[t]
    \centering
    \includegraphics[width=\linewidth]{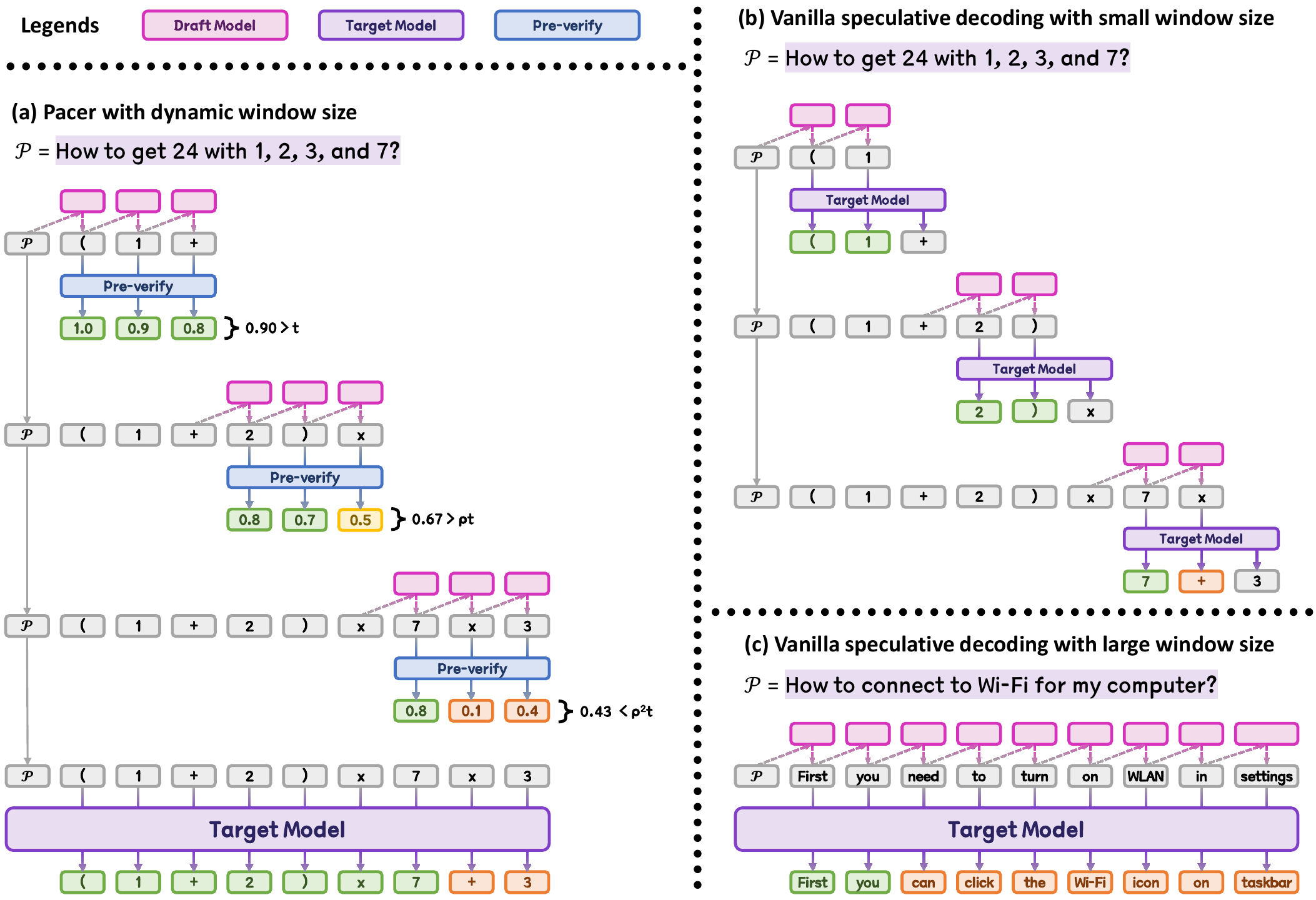}
    \caption{Comparison of the speculative decoding (SD) process for \method{} and vanilla SD with fixed small and large window sizes. (a) \method{} generates drafts in blocks of size \(b=3\), performs three rounds of pre-verification, and produces a total of \textbf{9} draft tokens, of which \textbf{7} are accepted with only \textbf{one} target model forward pass. (b) Vanilla SD with a small window size (\(\gamma=2\)) causes the draft model to stop prematurely, resulting in \textbf{3 costly} target forward and \textbf{6} draft forward. (c) Vanilla SD with a large window size \(\gamma=9\) generates \textbf{9} draft tokens, but only \textbf{2} are accepted, leading to wasted draft computation.}
    \label{fig:overview}
\end{figure}

Most existing approaches employ a fixed number of draft tokens, denoted as \(\gamma\)  (also referred to as the \textit{window size}), throughout the entire decoding process. This window size is typically determined manually by testing various values to achieve the best Speedup in experiments. However, we observe that the acceptance length varies significantly across different decoding steps, as illustrated in~\Cref{fig:accept_length}. This variation indicates that using a fixed window size often leads to suboptimal drafting behavior. The limitations of fixed window sizes are twofold: (1) overly conservative window size (small \(\gamma\)) results in unnecessary target model verification overhead (\Cref{fig:overview}(b)), while (2) excessively large windows (large \(\gamma\)) waste computation on ultimately rejected drafts (\Cref{fig:overview}(c)). Our preliminary experiments show that dynamically setting the optimal window size for each decoding step can improve overall throughput by up to \(1.4\times\) compared to using a fixed window size, as shown in \Cref{fig:optimal}. These findings highlight the need for an effective and efficient method to dynamically determine the optimal draft window size at each decoding step.

Several works have attempted to address this challenge by leveraging inherent outputs of draft models, such as token probabilities or entropy, to estimate token acceptance rates~\cite{gante2023assisted, brown_dynamicDDD_2024, wang2025opt, agrawal2024adaedl, zhang2024draft}. While these metrics reflect the model's semantic uncertainty to some extent, studies have shown that their accuracy often fluctuates with task and data distribution changes, making it difficult to ensure robust and consistent performance~\cite{valentin2024cost}. Additionally, LLMs are prone to miscalibration, particularly in the form of overconfidence~\cite{quevedo2024detecting, valentin2024cost}. Relying solely on these metrics overlooks the generated context and its alignment with the target model, leading to naive and unreliable acceptance rate estimates. On the other hand, PEARL~\cite{liu2025pearl} proposes a framework that runs the draft and target models in parallel. However, this approach struggles to efficiently schedule GPU resources in low-resource scenarios, particularly when both models are colocated on the same device. They compete for computational resources, ultimately slowing down execution. Additionally, PEARL requires the draft model’s total inference cost per step to be of the same order of magnitude as the target model’s, which limits its practicality.

To address this issue, we propose \method{}, a decoding framework that uses a trainable blockwise pre-verification module to dynamically regulate the draft window size during speculative decoding based on contextual information. Specifically, \method{} pre-verifies whether draft tokens are likely to be accepted before they are formally verified by the target model. As shown in \Cref{fig:overview}(a), the draft model first generates tokens in blocks of size \(b\), and a pre-verification layer is then introduced to pre-verify each block of \(b\) tokens. If the tokens in the current block pass the pre-verification, the draft model continues to generate the next \(b\)-token block; otherwise, it stops. This approach minimizes the misclassification of individual draft tokens and amortizes the additional inference overhead introduced by the pre-verification module. Furthermore, \method{} incorporates positional encoding for draft tokens, enabling more accurate predictions of acceptance probabilities and better optimization of dynamic window sizes.

We implement \method{} on multiple SD model pairs, including DeepSeek-Coder~\cite{guo2024deepseek}, Llama-2~\cite{touvron_llama_2023}, and Qwen-2.5~\cite{yang2024qwen2}, and evaluate its performance across various text generation benchmarks, such as code generation~\cite{Chen2021HumanEval}, mathematical reasoning~\cite{Cobbe2021TrainingVT}, and text summarization~\cite{see2017get}. Experimental results demonstrate that \method{} achieves up to \underline{\(2.66\times\) Speedup} over vanilla autoregressive decoding and consistently outperforms SD with a fixed window size. Moreover, when integrated with Ouroboros, \method{} achieves up to \underline{\(3.09\times\) Speedup} over autoregressive decoding. In summary, our main contributions are as follows:

\begin{itemize}
\item We present a systematic analysis of the performance gains enabled by adaptive draft lengths and provide key empirical insights that directly motivate the design of our dynamic draft-length framework.
\item We propose \method{}, an effective and efficient framework that dynamically controls and leads to an optimal draft window size by pre-verifying draft tokens blockwise and leveraging draft position information.
\item We conduct extensive evaluations of \method{} on multiple text generation benchmarks, demonstrating its effectiveness in consistently outperforming baseline methods and speculative decoding approaches with fixed window sizes.
\item We further show that \method{} is compatible with other speculative decoding methods designed to improve draft generation quality, and can be seamlessly integrated with such techniques to achieve additional performance gains, validating the universality of our approach.
\end{itemize}

\section{Preliminary}
\subsection{Notations}
In this paper, we denote the target model by \( M_T \), the draft model by \( M_D \), and our blockwise pre-verification layer by \( M_B \). Let \( \mathcal{V} \) represent the vocabulary set with size \( V \), and let \( \gamma \) denote the window size, i.e., the number of draft tokens generated by \( M_D \) before verification by \( M_T \). Given an input prefix sequence \( \mathbf{x} = [x_1, x_2, \dots, x_n] \) of length \( n \), we define the autoregressive generation of \( m \) tokens using model \( M \) as:
\begin{equation}
    (y_1, p_1), (y_2, p_2), \dots, (y_m, p_m) = M^m(\mathbf{x}),
\end{equation}
where \( y_1, \dots, y_m \in \mathcal{V} \) are the generated tokens, and \( p_1, \dots, p_m \in \mathbb{R}^V \) are their corresponding decoding distributions. Each draft token \( y_i \) is sampled according to its distribution, \( y_i \sim p_i \), for \( i = 1, \dots, m \).

For a parallel forward pass of model \( M \), we have:
\begin{equation}
    p_1, \dots, p_{m+1} = M(y_1, \dots, y_m; \mathbf{x}),
\end{equation}
where \( p_{m+1} \) denotes the distribution for the next token after \( y_m \).

\subsection{Speculative Decoding}
Speculative decoding consists of two stages: drafting and verification. Given an input prefix \( \mathbf{x} \), the draft model first autoregressively generates \( \gamma \) draft tokens:
\begin{equation}
    (y_1, q_1), \dots, (y_\gamma, q_\gamma) = M_D^\gamma(\mathbf{x}),
\end{equation}
where \( q_1, \dots, q_\gamma \) denote the candidate distributions. Subsequently, the target model verifies the drafts \( y_1, \dots, y_\gamma \) by performing a parallel forward pass with the concatenated prefix \( \mathbf{x} \):
\begin{equation}
    p_1, \dots, p_{\gamma + 1} = M_T(y_1, \dots, y_\gamma; \mathbf{x}).
\end{equation}
The acceptance rate of each draft token \( y_i \) is calculated as:
\begin{equation}
\alpha_i =
\begin{cases}
1, & \text{if } p_i[y_i] \geq q_i[y_i],\\[6pt]
\dfrac{p_i[y_i]}{q_i[y_i]}, & \text{otherwise}.
\end{cases}
\end{equation}
If token \( y_i \) is rejected, all subsequent tokens \( y_{i+1}, \dots, y_\gamma \) are discarded, and a new token is resampled from the normalized distribution \( \mathrm{norm}(\max(0, p_i - q_i)) \). If all drafts are accepted, SD samples an additional token from distribution \( p_{\gamma+1} \). This process allows SD to validate multiple tokens in one parallel step, substantially reducing the sequential steps required in autoregressive decoding. Importantly, the resulting output distribution remains consistent with that of vanilla autoregressive decoding of the target LLM~\cite{leviathan_fast_2023}.

\subsection{Acceptance Length}

The acceptance length \( L_A \) denotes the number of draft tokens accepted in a single SD step, which includes both the drafting and verification phases. It is bounded by the draft window size \( \gamma \), such that \( 0 \leq L_A \leq \gamma \). The value of \( L_A \) is jointly determined by the draft model and the target model, reflecting how well the draft model fits the target model's next-token distribution.

A larger fixed window size typically yields a longer acceptance length, but it also increases the number of draft tokens that may be rejected, leading to wasted draft-model computation. Conversely, a smaller fixed window size reduces draft waste but limits the attainable acceptance length, resulting in more frequent and costly target-model forward passes. These trade-offs highlight the importance of employing a dynamic window size that adapts to the acceptance behavior observed during decoding.

\section{Method}
In this section, we first present observations on acceptance lengths across decoding steps and demonstrate the speedup achieved by using optimal draft lengths. We then introduce \method{}, a decoding framework that dynamically controls the draft window size by pre-verifying draft tokens in a blockwise manner. Finally, we describe the training of \method{}.

\subsection{Observations}\label{sec:observation}

\begin{figure}[!h]
	\centering
	\subfigure[]{
		\includegraphics[width=0.48\linewidth,trim=0 0 0 0,clip]{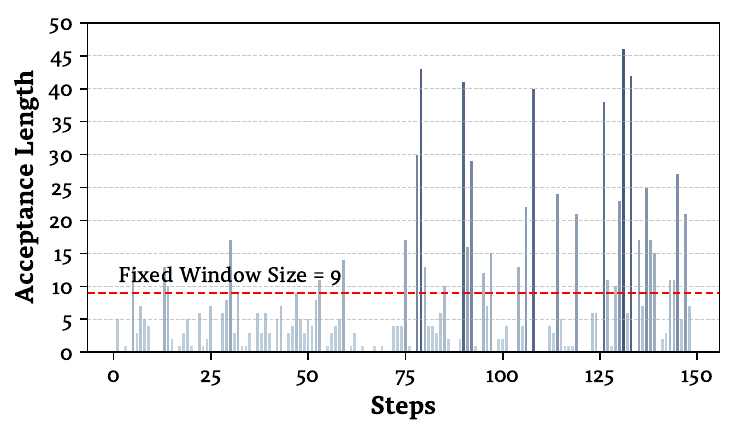}
        \label{fig:accept_length}
    }%
	\subfigure[]{
		\includegraphics[width=0.48\linewidth,trim=0 0 0 0,clip]{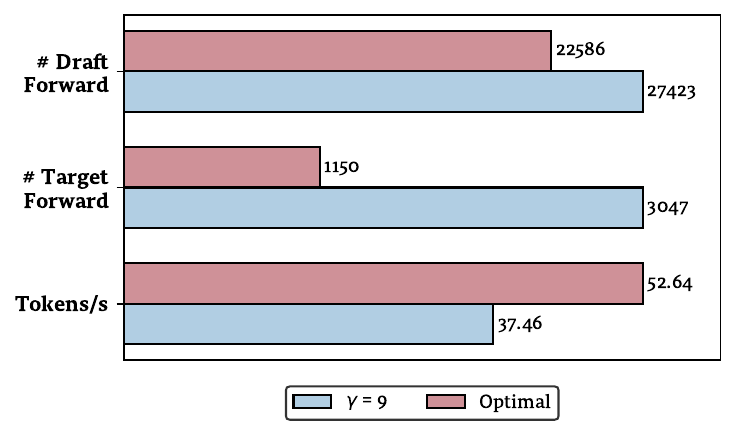}
        \label{fig:optimal}
    }%
    
    \caption{(a) Maximum acceptance lengths across decoding steps. The optimal fixed window size (\(\gamma=9\)) is marked by the red horizontal line. Significant variation in acceptance lengths highlights the inefficiency of employing a fixed draft window size. (b) Comparison between speculative decoding using the optimal fixed window (\(\gamma=9\)) and optimal dynamic window size (\(\gamma^\star\)). Utilizing dynamic draft lengths substantially reduces forward passes for both draft and target models, leading to increased decoding speed (tokens/s).}   
    
\vspace{-0.5cm}
\end{figure}

\subsubsection{Acceptance Lengths Vary Significantly Between Steps}
We conduct pilot experiments using DeepSeek-Coder models~\cite{guo2024deepseek}, employing the 1.3B model as the draft model and the 33B model as the target model on HumanEval~\cite{Chen2021HumanEval}. To obtain the maximum acceptance lengths \(L_A^\star\) , we set the draft window size sufficiently large during decoding. The acceptance lengths across different decoding steps are plotted in \Cref{fig:accept_length}.

Our analysis reveals significant variation in acceptance lengths between decoding steps, demonstrating the suboptimality of fixed window sizes. This manifests in two key inefficiencies:
\begin{enumerate}
    \item When \( \gamma > L_A^\star \), particularly when \( L_A^\star \approx 0 \), computation resources are wasted on generating ultimately rejected draft tokens.
    \item When \( \gamma < L_A^\star \) (the bars with long acceptance lengths in \Cref{fig:accept_length}), the draft model prematurely terminates, missing opportunities to generate additional accepted tokens. This results in unnecessary target model forward passes, which are computationally expensive.
\end{enumerate}
Thus, dynamically adjusting the window size is essential for optimizing the efficiency of speculative decoding.

\subsubsection{Optimal Draft Lengths Enhance SD Efficiency}
We define the optimal draft length \(\gamma^\star\) as the maximum acceptance length \(L_A^\star\) for each decoding step. As illustrated in \Cref{fig:optimal}, compared to the best-performing SD method with a fixed window size 9, SD using \(\gamma^\star\) reduces the number of draft forward passes by 4,837, and significantly decreases target forward passes from 3,047 to 1,150, resulting in an overall Speedup of \(1.4\times\). This improvement arises from minimizing computational overhead associated with redundant forward passes in both draft and target models. Thus, accurately estimating \(\gamma^\star\) can significantly enhance the efficiency of SD.

\subsubsection{Acceptance Rates Decrease with Draft Position}
\label{sec:draft_pos}
We analyze how the acceptance rate varies with the position of a draft token and visualize this trend in \Cref{fig:draft_pos}. The results show a sharp decline in acceptance rates as the draft position increases, indicating that tokens appearing later in the draft sequence are substantially less likely to be accepted by the target model. This suggests that it is possible to incorporate positional information when predicting acceptance probabilities.

\subsection{\method}
Based on the observations in \Cref{sec:observation}, we note that acceptance length strongly depends on the generated context. Therefore, we introduce \method, a trainable module designed to leverage draft contexts to approximately pre-verify draft tokens in a blockwise manner. The workflow of \method~is illustrated in \Cref{fig:overview}(a). To achieve a balance between prediction accuracy and computational overhead, \method~employs a single-layer Transformer atop the draft model to capture context and perform pre-verification predictions blockwise.

Our key insight is that accurately predicting draft acceptance rates requires considering all prior draft context. However, we observe that the forward-pass latency of the pre-verification layer is non-negligible. To balance accuracy and computational efficiency, we perform pre-verification in blocks of size \( b \), effectively amortizing this overhead. Specifically, given a prefix \(\mathbf{x}\), the draft model \( M_D \) begins by generating \( b \) draft tokens:
\begin{equation}
    (y_1,h_1),\dots,(y_b,h_b) = M_D^b(\mathbf{x}),
\end{equation}
where \( h_1,\dots,h_b \) represent the hidden states of the \( b \) drafts. The pre-verification layer \( M_B \) then processes these hidden states along with positional embeddings:
\begin{equation}
    \hat{\alpha}_1,\dots,\hat{\alpha}_b = M_B\left([h_1+e_1],\dots,[h_b+e_b];\mathbf{h_x}\right),
\end{equation}
Here, \( \hat{\alpha}_1,\dots,\hat{\alpha}_b \) represent the estimated acceptance rates for drafts \( y_1,\dots,y_b \), while \( e_1,\dots,e_b \) denote positional embeddings corresponding to draft positions \( 1 \) to \( b \). Additionally, \( \mathbf{h_x} \) represents the hidden states of previously accepted tokens before the current draft-and-verify step. We compute the mean estimated acceptance rate across the block, \( \text{mean}(\hat{\alpha}_1,\dots,\hat{\alpha}_b) \), and stop drafting if this mean falls below a predefined threshold \( t \), triggering verification by the target model \( M_T \). Otherwise, \( M_D \) continues to generate another block of \( b \) tokens. For next pre-verification rounds, the draft positions become \( b+1,\dots,2b \), and so forth. For drafts with \(k\) pre-verification rounds, the final window size for that decoding step is \(\gamma = k \cdot b\).  Meanwhile, the threshold \( t \) is increased each round by a growth factor \(\rho>1\), making it progressively easier to stop draft generation in later steps. The detailed algorithm is presented in~\Cref{sec:algorithm}.

\subsection{Training}\label{sec:training}
To ensure consistency between training and inference, we first utilize the target model \( M_T \) to generate responses based on instructions from the training datasets. We then apply the same speculative decoding process used during inference to construct labeled acceptance data. Specifically, the draft model \( M_D \) starts from a given prefix sequence \(\mathbf{x}\) and generates a sequence of draft tokens \( y_1, \dots, y_\gamma \) using a large window size (e.g., \(\gamma = 50\)). We compare these drafts against tokens generated by the target model: if a draft token \( y_i \) differs from the corresponding target token, all subsequent tokens \( y_i, \dots, y_\gamma \) are labeled as 0 (rejected), while tokens \( y_1, \dots, y_{i-1} \) are labeled as 1 (accepted). This procedure produces labeled training data aligned precisely with the inference-time verification process. We train the pre-verification layer \(M_B\) using standard cross-entropy loss. To improve training efficiency, we pack draft tokens from multiple decoding steps into a single sequence with carefully designed attention masks. Further details about training are provided in~\Cref{sec:training_detail}.

\section{Experiments}\label{sec:experiments}
\subsection{Experiments Setting}
\paragraph{Evaluation Datasets.}
We evaluate \method{} on multiple representative text-generation benchmarks to demonstrate its effectiveness. Specifically, for code generation tasks, we utilize two widely adopted benchmarks: HumanEval~\citep{Chen2021HumanEval} and MBPP~\citep{austin2021mbpp}. For document summarization, we employ the well-established CNN/Daily~Mail (CNN/DM) dataset~\citep{see2017get}. Additionally, for arithmetic reasoning, we use the GSM8K dataset~\citep{Cobbe2021TrainingVT}. Further details on evaluation datasets can be found in \Cref{sec:eval_setting}.

\paragraph{Training Datasets.}
We use the CodeAlpaca dataset~\citep{codealpaca}, which contains approximately 20,000 instruction-response pairs covering a wide range of programming scenarios and coding tasks. We use the instructions to generate our training data.

\paragraph{Models.}
We employ several state-of-the-art LLMs widely used in current research, including the DeepSeek-Coder series~\citep{guo2024deepseek}, the Llama2 series~\citep{touvron_llama_2023}, and the Qwen2.5 series~\citep{yang2024qwen2}. Specifically, we adopt \textbf{1.3B/33B} and \textbf{6.7B/33B} models from the DeepSeek-Coder series, \textbf{7B/70B} models from the Llama2-chat series, and \textbf{1.5B/32B} models from the Qwen2.5 series, using less-parameter models as drafts and their larger counterparts as target models. These models encompass a wide range of model sizes representative of LLM research.

\paragraph{Evaluation Methods.}
We compare our method against several established baselines, including vanilla autoregressive decoding, speculative decoding~\citep{leviathan_fast_2023, chen_accelerating_2023}, lookahead decoding~\citep{fu2024break}, a heuristic-based dynamic window method (assist generation)~\citep{gante2023assisted}, and retrieval-based decoding (REST)~\citep{he_rest_2024}. To ensure fair comparisons on our hardware setup, we reproduced each baseline using their official implementations and default parameters. In all experiments, including both baselines and our proposed method, we set the batch size to 1, which is standard across speculative decoding frameworks. We evaluate model performance using the following metrics: \textbf{decoding speed (tokens/s)}, \textbf{speedup ratio}, and \textbf{average acceptance lengths (\(\tau\))}. The detailed evaluation settings can be found in \Cref{sec:eval_setting}.

\begin{figure}[t]
    \centering
    \includegraphics[width=1\linewidth]{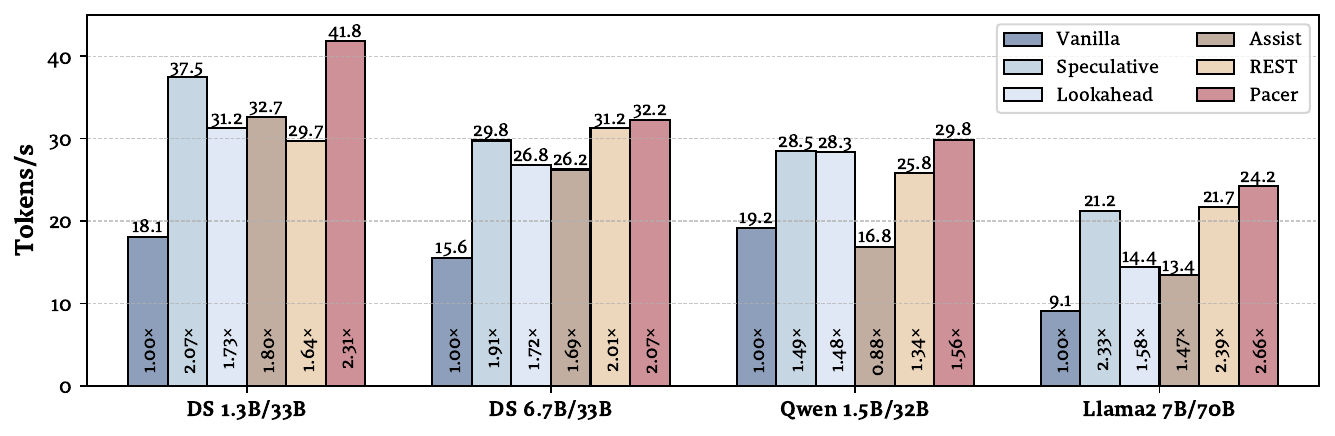}
\vspace{-0.5cm}
    \caption{Comparison of decoding speeds (tokens/s) and speedup for different methods on the HumanEval dataset. \method{} consistently outperforms all baseline methods across various models.}
    \label{fig:human_eval}
\end{figure}

\begin{table}[htbp]
\centering
    \renewcommand{\arraystretch}{1.0}
\small
\begin{tabular}{lcccccc}
\toprule
\multirow{2.5}{*}{\textbf{Algorithm}} & \multicolumn{2}{c}{\textbf{Deepseek 1.3B/33B}} & \multicolumn{2}{c}{\textbf{Deepseek 6.7B/33B}} & \multicolumn{2}{c}{\textbf{Llama-2 7B/70B}} \\
\cmidrule(lr){2-3} \cmidrule(lr){4-5} \cmidrule(lr){6-7}
& tokens/s & speedup & tokens/s & speedup & tokens/s & speedup \\
\midrule
\rowcolor{gray!15} \multicolumn{7}{c}{\textit{MBPP}} \\
Vanilla      & 17.53 & 1.00 & 15.86 & 1.00 &  9.28 & 1.00 \\
Speculative  & 30.72 & 1.75 & 26.62 & 1.68 & 18.65 & 2.01 \\
Lookahead    & 29.55 & 1.69 & 23.67 & 1.49 & 13.30 & 1.43 \\
Assist       & 26.60 & 1.52 & 21.07 & 1.33 & 16.30 & 1.76 \\
REST         & 29.41 & 1.69 & 25.64 & 1.60 & 15.38 & 1.66 \\
\method        & \textbf{32.93} & \textbf{1.88} & \textbf{28.57} & \textbf{1.80} & \textbf{19.90} & \textbf{2.14} \\
\rowcolor{gray!15} \multicolumn{7}{c}{\textit{CNN/DM}} \\
Vanilla      & 17.11 & 1.00 & 15.37 & 1.00 &  8.46 & 1.00 \\
Speculative  & 23.55 & 1.38 & 20.55 & 1.34 & 14.40 & 1.70 \\
Lookahead    & 18.21 & 1.08 & 17.64 & 1.15 &  8.56 & 1.01 \\
Assist       & 22.27 & 1.30 & 18.52 & 1.20 & 11.36 & 1.34 \\
REST         & 18.50 & 1.06 & 16.70 & 1.09 &  9.62 & 1.14 \\
\method        & \textbf{23.77} & \textbf{1.39} & \textbf{21.51} & \textbf{1.40} &   \textbf{14.50}  &  \textbf{1.71}  \\
\rowcolor{gray!15} \multicolumn{7}{c}{\textit{GSM8k}} \\
Vanilla      & 18.25 & 1.00 & 16.51 & 1.00 &  9.19 & 1.00 \\
Speculative  & 35.08 & 1.92 & 29.36 & 1.78 & 17.81 & 1.94 \\
Lookahead    & 33.82 & 1.85 & 29.09 & 1.76 & 12.07 & 1.31 \\
Assist       & 34.86 & 1.91 & 28.80 & 1.74 & 14.02 & 1.53 \\
REST         & 23.25 & 1.27 & 22.22 & 1.34 & 12.66 & 1.38 \\
\method        & \textbf{39.69} & \textbf{2.17} & \textbf{31.07} & \textbf{1.88} & \textbf{19.07} & \textbf{2.08} \\
\bottomrule
\end{tabular}
\caption{Decoding speed (tokens/s) and speedup across different methods and LLM families on MBPP, CNN/DM and GSM8K benchmarks.}
\label{tab:main_result}
\end{table}
\subsection{Main Results}
We conducted experiments on the benchmarks described previously. As shown in \Cref{fig:human_eval}, \method{} consistently outperformed vanilla autoregressive decoding, speculative decoding, lookahead decoding, assist generation, and REST across all evaluated model series on the HumanEval dataset. Specifically, in experiments using DeepSeek-Coder 1.3B/33B, \method{} achieved a decoding speed of \(41.8\) tokens/s, corresponding to a \(2.31\times\) speedup over vanilla autoregressive decoding and surpassing the \(2.07\times\) speedup achieved by vanilla speculative decoding. For Llama-2 7B/70B, the speedup improved from \(2.33\times\) to \(2.66\times\). \Cref{tab:main_result} further presents decoding speeds and corresponding speedup ratios for various methods across multiple datasets. \method{} consistently achieved the highest decoding speeds and speedups across all models, outperforming all tested methods. Notably, on the GSM8K dataset, the speedup ratio improved from \(1.92\times\) to \(2.17\times\) with DeepSeek-Coder 1.3B/33B and from \(1.94\times\) to \(2.08\times\) with Llama-2 7B/70B. These results demonstrate that utilizing a blockwise pre-verification layer to dynamically control draft lengths effectively leverages the capabilities of the draft model, yielding significant performance improvements across diverse tasks and model setups.

\subsection{Comparison with Other Dynamic Draft Length Methods}\label{comaprison of other dynaminc methods}

\begin{table*}[ht]
\centering
\small

\setlength\tabcolsep{3.5pt}
\begin{tabular}{lccccccc}
\toprule
\textbf{Method} & \textbf{MT-Bench} & \textbf{Translation} & \textbf{Summarization} & \textbf{QA} & \textbf{Math Reasoning} & \textbf{RAG} & \textbf{Average} \\
\midrule
Vanilla      & 9.04  & 9.31  & 8.63  & 9.32  & 9.29  & 8.32  & 8.99  \\
Assist       & 17.77 & 17.95 & 16.42 & 17.10 & 20.46 & 15.60 & 17.58 \\
AdaEDL       & 17.55 & 17.85 & 16.10 & 17.34 & 20.21 & 15.69 & 17.47 \\
SpecDec++    & 18.18 & 17.62 & \textbf{16.57} & 17.41 & 20.81 & \textbf{15.80} & 17.79 \\
\method{} & \textbf{18.53} & \textbf{18.06} & 16.19 & \textbf{17.42} & \textbf{21.45} & 15.48 & \textbf{17.95} \\
\bottomrule
\end{tabular}
\caption{Decoding speed (tokens/s) on SpecBench with Llama-2 7B/70B.}
\label{tab:specbench}
\end{table*}

\begin{table}[h]
\centering
\small
\begin{tabular}{lcc}
\toprule
\textbf{Method}      & \textbf{tokens/s} & \textbf{$\tau$} \\
\midrule
Speculative                   & 21.20             & 5.39                        \\
SpecDec++$^*$            & 21.17             & 5.13                        \\
AdaEDL               & 22.53             & 4.57                        \\
\method{}      & \textbf{24.20}    & \textbf{7.46}               \\
\bottomrule
\end{tabular}
\caption{Decoding speed (tokens/s) and average acceptance length ($\tau$) of dynamic draft length methods on HumanEval with Llama-2 7B/70B. $^*$SpecDec++ underperforms standard SD here due to differences in chat templates compared to its original paper. We present results obtained using the original SpecDec++ templates in the appendix, where \method{} remains superior.}
\label{tab:dynamic_draft_comparison_humaneval}

\end{table}

We benchmark \method{} against several state-of-the-art methods that utilize dynamic draft lengths, including AdaEDL~\citep{agrawal2024adaedl}, which generates drafts based on the entropy of draft token distributions, and SpecDec++~\citep{huang_specdec_2024}, which controls draft lengths using an additional prediction head. We conduct a focused comparison on the HumanEval benchmark. As shown in \Cref{tab:dynamic_draft_comparison_humaneval}, \method{} outperforms all other dynamic methods, achieving the highest decoding speed (24.20 tokens/s) and significantly longer average acceptance length (7.46). These results highlight the effectiveness of our pre-verification layer.

Additionally, we evaluate these methods on the SpecBench benchmark~\citep{xia-etal-2024-unlocking}. As shown in \Cref{tab:specbench}, \method{} consistently achieves the highest average decoding speed, confirming the generalizability of our pre-verification mechanism.

\subsection{Integration with Ouroboros}\label{integration with ouroboros}
To demonstrate the flexibility and orthogonality of \method{}, we integrated it with Ouroboros~\citep{zhao2024ouroboros}, a recent method that improves drafting efficiency by generating phrase-level drafts from an n-gram pool. While Ouroboros enhances draft quality, it employs a fixed draft steps. In contrast, \method{} dynamically adjusts draft lengths via pre-verification. We hypothesized that combining these complementary strategies could yield further performance improvements. We conducted experiments on the HumanEval, MBPP, and GSM8K benchmarks using DeepSeek-Coder 1.3B/33B, using the optimal configuration from the Ouroboros paper.

\begin{table}[htbp]
\centering
\small
\setlength\tabcolsep{4pt}
\begin{tabular}{llccc}
\toprule
\textbf{Dataset} & \textbf{Method} & \textbf{tokens/s} & \textbf{speedup} & $\tau$ \\
\midrule
\multirow{3}{*}{\textit{HumanEval}} & Vanilla & 18.10 & 1.00 & — \\
& Ouroboros & 49.05 & 2.71 & 8.36 \\
& \method{} + Ouroboros & \textbf{51.07} & \textbf{2.82} & \textbf{10.90} \\
\midrule
\multirow{3}{*}{\textit{MBPP}} & Vanilla & 17.53 & 1.00 & — \\
& Ouroboros & 47.99 & 2.74 & 3.81 \\
& \method{} + Ouroboros & \textbf{50.08} & \textbf{2.86} & \textbf{5.10} \\
\midrule
\multirow{3}{*}{\textit{GSM8K}} & Vanilla & 18.25 & 1.00 & — \\
& Ouroboros & 52.66 & 2.89 & 4.43 \\
& \method{} + Ouroboros & \textbf{56.31} & \textbf{3.09} & \textbf{5.99} \\
\bottomrule
\end{tabular}
\caption{Comparison of decoding speed (tokens/s) and the average acceptance lengths ($\tau$) on benchmark HumanEval, MBPP, and GSM8K using Deepseek-Coder 1.3B/33B, when combining \method{} with Ouroboros.}
\label{tab:pacer_ouroboros}
\end{table}

As shown in \Cref{tab:pacer_ouroboros}, the integrated approach consistently achieves higher decoding speeds and longer average acceptance lengths. For example, on GSM8K, the speedup increases from \(2.89\times\) with Ouroboros alone to \(3.09\times\) with the combined method. These findings highlight the adaptability of \method{} and its effectiveness in enhancing other SD methods that optimize the draft process.

\subsection{Ablation Studies}
\begin{table}[h!]
\centering
\small
\begin{tabular}{lcccccc}
\toprule
\multirow{2.5}{*}{\textbf{Method}} & \multicolumn{3}{c}{\textbf{HumanEval}} & \multicolumn{3}{c}{\textbf{GSM8K}} \\
\cmidrule(lr){2-4} \cmidrule(lr){5-7}
& tokens/s & speedup & \(\tau\) & tokens/s & speedup & \(\tau\) \\
\midrule
Vanilla               & 18.11 & 1.00 & --    & 18.25 & 1.00 & --    \\
Speculative   & 37.46 & 2.07 & 7.38  & 35.08 & 1.94 & 5.87  \\
\midrule
\method                    & \textbf{41.80} & \textbf{2.31} & 9.67  & \textbf{39.69} & \textbf{2.17} & 7.23  \\
\quad w/o pos-emb    & 39.99 & 2.21 & 8.83  & 37.48 & 2.05 & 7.19  \\
\quad w/o growth-factor  & 40.36 & 2.23 & \textbf{9.71}  & 37.84 & 2.07 & \textbf{7.46}  \\
\bottomrule
\end{tabular}
\caption{Ablations on draft position embedding and threshold growth factor with DeepSeek-Coder 1.3B/33B.}
\label{tab:ablation}
\end{table}

We conduct a series of ablation studies to examine the key components of our approach. Specifically, we analyze the impact of draft position embeddings (\textbf{pos-emb}) and the threshold growth factor (\textbf{growth-factor}) on overall performance. We also carried out additional ablations to analyze our hyperparameter settings and key aspects of our model design. Additional ablations are presented in appendix.

\textbf{Effect of Position Embedding.} As shown in \Cref{tab:ablation}, removing draft position embeddings reduces the decoding speed to \(39.99\) tokens/s, indicating that position information contributes significantly to the overall performance of \method{}. This aligns with the observation that draft position serves as a strong indicator of acceptance rate, and incorporating it enhances prediction accuracy.

\textbf{Effect of Growth Factor.} As shown in \Cref{tab:ablation}, removing the threshold growth factor increases the average acceptance length \(\tau\), but results in lower decoding speed. This indicates that the growth factor plays a crucial role by progressively increasing the likelihood of early stopping as draft sequences grow longer. Without it, \method{} tends to generate overly long drafts, leading to wasted computation and reduced efficiency.

\section{Discussion}
In this section, we investigate the underlying factors that enable \method{} to accelerate speculative decoding and discuss key architectural design choices. We analyze the performance contributions of \method{}, examine alternative halting criteria for draft generation, and evaluate the impact of different attention scopes in the pre-verification layer. 

\subsection{Where Does the Speedup of \method{} Come From?}

To understand the performance advantages of \method{}, we record the number of forward passes for both the draft and target models, along with the average acceptance length. As shown in \Cref{tab:num_forward}, \method{} with DeepSeek-Coder 1.3B/33B achieves substantial improvements over speculative decoding with the optimal fixed window size across HumanEval, CNN/DM, and GSM8K. \method{} reduces the forward passes for both the draft and target models while simultaneously increasing the average acceptance length.

These improvements arise from \method{}'s ability to adaptively adjust the draft window size. It halts early in regions where predictions are uncertain, preventing wasted draft computation, and produces longer drafts in segments where acceptance likelihood is high. This adaptive behavior improves data-movement efficiency, alleviates the memory bound inherent in autoregressive decoding, and yields consistent speed gains across tasks and model scales.

\begin{table}[h!]
\centering
\small
 \setlength\tabcolsep{5pt}
\begin{tabular}{lccccccccc}
\toprule
\multirow{3.5}{*}{\textbf{Methods}} & \multicolumn{3}{c}{\textbf{HumanEval}} & \multicolumn{3}{c}{\textbf{CNN/DM}} & \multicolumn{3}{c}{\textbf{GSM8k}} \\
\cmidrule(lr){2-4} \cmidrule(lr){5-7} \cmidrule(lr){8-10}
& \makecell{\#Draft\\Forward} & \makecell{\#Target\\Forward} & \makecell{\(\tau\)} & \makecell{\#Draft\\Forward} & \makecell{\#Target\\Forward} & \makecell{\(\tau\)} & \makecell{\#Draft\\Forward} & \makecell{\#Target\\Forward} & \makecell{\(\tau\)} \\
\midrule
Speculative & 27423 & 3047 & 7.38 & 23904 & 7968 & 1.78 & 59456 & 7432 & 5.87 \\
\method         & \textbf{26376} & \textbf{2368} & \textbf{9.67} & \textbf{22252} & \textbf{7884} & \textbf{1.89} & \textbf{54024} & \textbf{6816} & \textbf{7.23} \\
\bottomrule
\end{tabular}
\caption{Comparison of the number of forward passes for the draft and target models, and the average acceptance lengths (\(\tau\)), between vanilla speculative decoding with optimal fixed window size (Speculative) and \method{} across different benchmarks.}
\label{tab:num_forward}
\end{table}

\subsection{What Is the Best Halting Criterion for Draft Generation?}
To determine an effective halting criterion for the pre-verification stage, we compare our proposed mean-token probability criterion with two stricter alternatives: (1)~\textbf{Any-token below $t$}: stopping when \textbf{any} token in the block has predicted acceptance below threshold~$t$, and (2)~\textbf{Last-token below $t$}: stopping based solely on the predicted acceptance of the \textbf{last} token in the block. Results are shown in \Cref{tab:halting_criterion}.

\begin{table}[H]
    \centering
    \small
    \setlength\tabcolsep{10pt}
    \caption{Comparison of decoding speed and average acceptance length~($\tau$) using different halting criteria on HumanEval with DeepSeek-Coder 1.3B/33B.}
    \label{tab:halting_criterion}

    \begin{tabular}{lcc}
        \toprule
        \textbf{Halting Criterion} & \textbf{Tokens/s} & \textbf{$\tau$} \\
        \midrule
        Any-token below $t$ & 38.96 & 8.70 \\
        Last-token below $t$ & 40.11 & 9.34 \\
        Mean-token probability $t$ (\textbf{Ours}) & \textbf{41.80} & \textbf{9.67} \\
        \bottomrule
    \end{tabular}
\end{table}

As seen in \Cref{tab:halting_criterion}, our mean-token criterion achieves the highest decoding speed and the longest accepted sequence length. This is because averaging across tokens mitigates the effects of occasional low-confidence predictions. In contrast, stricter criteria tend to halt the draft prematurely, reducing acceptance lengths and lowering overall throughput.

\subsection{How Does Attention Context Affect Pre-verification?}
To validate the importance of full-context attention in the pre-verification layer, we compare it with two restricted variants: (1) \textbf{Local block}, where attention is limited to the current block, and (2) \textbf{Local draft}, where attention is restricted to the current draft step. Results are summarized in \Cref{tab:attention_context}.

\begin{table}[H]
    \centering
    \small
    \setlength\tabcolsep{10pt}
    \caption{Ablation on Attention Context}
    \label{tab:attention_context}
    \begin{tabular}{lcc}
        \toprule
        \textbf{Context Scope} & \textbf{Tokens/s} & \textbf{$\tau$}\\
        \midrule
        Local block & 22.29 & 5.32 \\
        Local draft & 22.58 & 5.45 \\
        Full context (\textbf{Ours}) & \textbf{24.20} & \textbf{7.46} \\
        \bottomrule
    \end{tabular}
\end{table}

Expanding the attention scope consistently improves decoding speed and acceptance length~$\tau$. Full-context attention is particularly effective because it captures long-range dependencies within the draft and prefix, enabling more reliable acceptance prediction.

\section{Related Work}
\paragraph{Speculative Decoding}
Speculative decoding (SD) utilizes a draft-verify paradigm to achieve lossless acceleration. Existing SD frameworks can generally be categorized based on the type of draft models used~\citep{xia_SDsurvey_2024}. (1) \textbf{Independent draft models}: SpecDec~\citep{xia-2023-speculative} introduced a non-autoregressive (non-AR) Transformer that generates multiple tokens simultaneously. Several other works~\citep{leviathan_fast_2023, chen_accelerating_2023, spector_accelerating_2023, Sun_SpecTr_2023} propose leveraging smaller pretrained models from the same LLM family (e.g., Llama2-7B and Llama2-70B~\citep{touvron_llama_2023}) for inference acceleration, thus avoiding additional training and maintaining alignment in prediction behaviors due to shared tokenizers and training corpora. SpecInfer~\citep{miao2024specinfer} and DistillSpec~\citep{zhou2024distillspec} adopt distillation techniques to obtain smaller draft models. (2) \textbf{Self-drafting}: This approach eliminates the need for external draft models by employing the target LLM itself for drafting, thereby ensuring close alignment. \citet{Stern_Blockwise_2018, cai_medusa_2024, hwang2024hydra, gloeckle2024better} train multiple heads to simultaneously predict multiple future draft tokens. Eagle~\citep{li_eagle_2024} introduces a single-layer transformer to perform autoregressive prediction at the feature level. Additionally, non-autoregressive decoding methods for generating drafts have been explored by \citet{santilli-2023-accelerating, yi-etal-2024-generation, xiao2024parallelspec}. Lookahead~\citep{fu2024break} constructs n-gram pools using Jacobi iterations as drafts. Meanwhile, works by \citet{yang_predictive_2024, zhang_draft_2024, elhoushi2024layerskip} investigate early exiting and layer skipping within the target model to enhance drafting efficiency. Despite these developments, most approaches utilize fixed window sizes, leaving the challenge of adaptive draft generation largely unaddressed.

\paragraph{Adaptive Draft Length}
Several recent studies have explored the adaptive adjustment of draft lengths. Works by \citet{zhang_draft_2024, liu_kangaroo_2024, li_eagle-2_2024, gante2023assisted, brown_dynamicDDD_2024, wang2025opt, agrawal2024adaedl, zhang2024draft} propose determining draft-stopping criteria using intrinsic outputs from draft models, such as token probability or entropy, to estimate draft acceptance rates. These methods typically rely on manually set thresholds or update rules to establish when to stop drafting. Although heuristic-based approaches offer simplicity and computational efficiency, relying solely on semantic uncertainty metrics lacks robustness and consistent accuracy. This is primarily due to the inherent miscalibration and overconfidence exhibited by LLMs~\citep{quevedo2024detecting, valentin2024cost, huo2025c2t}, resulting in overly simplistic and unreliable estimations of acceptance rates. SpecDec++~\citep{huang_specdec_2024} and DISCO~\citep{mamou2024dynamic} address this by training classifiers on top of draft models to determine when to stop token generation, but they neglect draft-context information. Meanwhile, PEARL~\citep{liu2025pearl} introduces a parallel framework allowing concurrent operation of target and draft models; however, it faces challenges handling resource constraints due to competition when both models are colocated. To address these limitations, we propose \method, a decoding framework that leverages a trainable blockwise pre-verification layer incorporating contextual information to efficiently and accurately determine the optimal draft length.

\section{Conclusion}
In this paper, we propose \method{}, an effective and efficient approach that dynamically controls draft lengths based on contextual information. \method{} employs a trainable pre-verification layer to blockwise pre-verify draft tokens. The draft model continues generating tokens only when the block passes pre-verification. This adaptive strategy effectively mitigates computational waste associated with fixed window sizes in speculative decoding. We conduct comprehensive experiments across diverse benchmarks, and \method{} consistently outperforms all speculative decoding baselines across model architectures, demonstrating its effectiveness in handling variable draft lengths.

\bibliography{colm2025_conference}
\bibliographystyle{colm2025_conference}

\newpage
\appendix

\section{Decoding Algorithm}\label{sec:algorithm}
\begin{algorithm}[ht]
\caption{Speculative Decoding with \method}
\begin{algorithmic}[1]
\Require Draft model $M_D$, target model $M_T$, pre-verification layer $M_B$, input prefix $\mathbf{x}$, maximum token length $L$, block size $b$, initial threshold $t$, growth factor $\rho$
\While{$\text{len}(\mathbf{x}) < L$}
    \State \textcolor{codeblue}{$\triangleright$ Adaptive draft generation}
    \State $\mathbf{y}, \mathbf{q} \gets [], []$ \quad \textcolor{codepurple}{$\triangleright$ 
 Draft tokens and Draft probabilities}
    \State $i \gets 1$
    \While{True}
        \State $(y_1, q_1, h_1), \dots, (y_b, q_b, h_b) \gets M_D^b(\mathbf{x})$
        \State $e_{1}, \dots, e_{b} \gets \mathrm{PosEmb}(i, \dots, i + b - 1)$
        \State \textcolor{codeblue}{$\triangleright$ Blockwise pre-verification with past KV cache}
        \State $\hat{\alpha}_1, \dots, \hat{\alpha}_b \gets M_B([h_1 + e_1], \dots, [h_b + e_b])$
        \State $i \gets i + b$
        \State $\mathbf{x} \gets \mathbf{x} + [y_1, \dots, y_b]$
        \State $\mathbf{y}$.append($[y_1, \dots, y_b]$)
        \State $\mathbf{q}$.append($[q_1, \dots, q_b]$)
        \If{$\text{mean}(\hat{\alpha}_1, \dots, \hat{\alpha}_b) \le t$}
            \State \textbf{break}
        \EndIf
        \State $t \gets t \cdot \rho$
    \EndWhile
    \State $\mathbf{x}, \mathbf{y} \gets \mathbf{x}$ \quad \textcolor{codepurple}{$\triangleright$ Split to get original input prefix }
    \State $y_1, \dots, y_\gamma \gets \mathbf{y}$
    \State $q_1, \dots, q_\gamma \gets \mathbf{q}$
    \State \textcolor{codeblue}{$\triangleright$ Target model verification}
    \State $p_1, \dots, p_{\gamma + 1} \gets M_T(y_1, \dots, y_\gamma; \mathbf{x})$
    \State Sample $r_1, \dots, r_\gamma \sim U(0,1)$
    \State $n \gets \min\left( \{ i - 1 \mid 1 \le i \le \gamma, r_i > \frac{p_i(y_i)}{q_i(y_i)} \} \cup \{\gamma\}\right)$
    \State \textcolor{codeblue}{$\triangleright$ Adjust target distribution if drafts are rejected}
    \If{$n < \gamma$}
        \State \textcolor{codered}{$\times$ reject draft token $y_n$}
        \State $p' \gets \text{norm}(\max(0, p_{n+1} - q_{n+1}))$
    \Else
        \State \textcolor{codegreen}{$\checkmark$ accept all draft tokens}
        \State $p' \gets p_{n+1}$
    \EndIf
    \State Sample next token $t \sim p'$
    \State $\mathbf{x} \gets \mathbf{x} + [y_1, \dots, y_n, t]$
\EndWhile
\end{algorithmic}
\end{algorithm}

\section{Case Study}
\begin{figure}[h!]
    \centering
    \includegraphics[width=\linewidth]{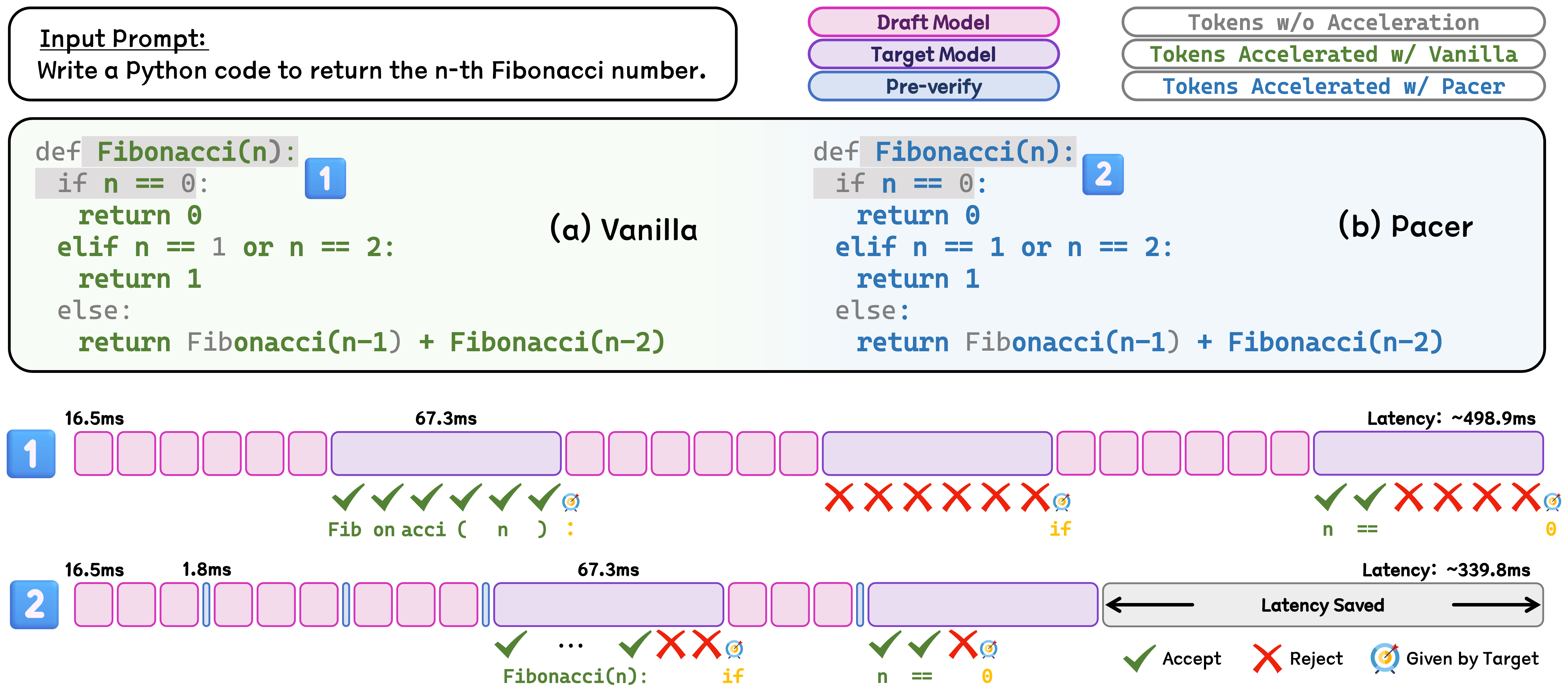}
    \caption{An illustrative comparison between vanilla SD with a fixed window size (\(\gamma = 6\)) and \method{} with a block size of \(b = 3\). The top part of the figure shows the accelerated token generation achieved by each method. The bottom part presents a latency breakdown for generating the code snippet \colorbox{backgray}{\textcolor{textgray}{\textbf{\texttt{Fibonacci(n): if n == 0}}}}.}
    \label{fig:case_study}
\end{figure}

we provide an illustrative example comparing \method{} with standard speculative decoding, as shown in \Cref{fig:case_study}. In this example, configuration (a) uses a fixed window size of $\gamma = 6$, whereas configuration (b) applies \method{} with a dynamic draft window and a block size of $b = 3$. Under the fixed-window setting, all six draft tokens are accepted in the first round; however, in the second round, all six draft tokens are rejected, resulting in unnecessary draft computation and requiring three target-model forward passes in total. In contrast, \method{} halts early when draft predictions become unreliable, avoids wasted drafts, and completes the decoding process with only two target forward passes. The latency breakdown clearly demonstrates the benefit of adaptive draft control: \method{} reduces total decoding time from 498.9 ms to 339.8 ms in this example, highlighting its effectiveness in lowering computational overhead and improving end-to-end efficiency.

\section{Experimental Details}
\subsection{Evaluation Settings}\label{sec:eval_setting}
\paragraph{Datasets.}
We evaluate our method on a range of representative text generation tasks, including HumanEval~\citep{Chen2021HumanEval} and MBPP~\citep{austin2021mbpp} for code generation, GSM8K~\citep{Cobbe2021TrainingVT} for arithmetic reasoning, and CNN/DailyMail (CNN/DM)~\citep{see2017get} for document summarization. HumanEval comprises 164 handcrafted examples, each consisting of a text prompt and a Python function prefix. MBPP originally contains 500 diverse programming problems with corresponding test cases; we randomly sample 150 problems for evaluation. The CNN/DM dataset includes news articles paired with human-written summaries, while GSM8K features grade-school-level mathematical word problems. For both CNN/DM and GSM8K, we randomly sample 100 examples for evaluation, following~\citep{zhao2024ouroboros}.

\paragraph{Hardware Setting.}
To validate the effectiveness of our method under both resource-constrained and resource-abundant conditions. The experiments of deepseek-coder 1.3B/33B and Qwen-2.5 1.5B/32B are performed on \textbf{1 × NVIDIA 80GB A800 GPU}, deepseek-coder 6.7B/33B on \textbf{2 × NVIDIA 80GB A800 GPU} and Llama2-chat 7B/70B on \textbf{4 × NVIDIA 80GB A800 GPU}.

\paragraph{Speculative Decoding.} In \Cref{sec:experiments}, we report the performance of vanilla speculative decoding using the best decoding speed achieved across various fixed window sizes. The complete results are provided in \Cref{sec:sd_results}.

\subsection{Training Parameters}

For each model series, we independently train a blockwise pre-verification layer. We use the AdamW optimizer with \((\beta_1, \beta_2) = (0.9, 0.999)\). All training is performed on \textbf{8$\times$ NVIDIA A800 GPUs}. Table~\ref{tab:training-param} summarizes the model scales and training configurations. The training time ranged from 18 to 47 minutes, depending on the model size.

\begin{table}[H]
    \small
    \centering
    \begin{tabular}{lrrrr}
        \toprule
        \textbf{Model Series} & \textbf{Pre-verification Layer Size} & \textbf{Learning Rate} & \textbf{Epochs} \\
        \midrule 
        DeepSeek-Coder 1.3B/33B    & 0.18B & 5e-4 & 5 \\
        DeepSeek-Coder 6.6B/33B    & 0.47B & 2e-4 & 5 \\
        Qwen2.5 1.5B/32B           & 0.28B & 5e-4 & 5 \\
        LLama2-Chat 7B/70B         & 0.46B & 1e-4 & 5 \\
        \bottomrule
    \end{tabular}
    \caption{Training hyperparameters for \method{} across different model series.}
    \label{tab:training-param}
\end{table}

\subsection{Evaluation Parameters}
The hyperparameters of our experiments are shown in \cref{tab:eval_params}.

\begin{table}[h!]
\centering
\small
\begin{tabular}{llccc}
\toprule
\textbf{Task} & \textbf{Model} & \textbf{Block Size \(b\)} & \textbf{Threshold \(t\)} & \textbf{Growth Factor \(\rho\)} \\
\midrule
\multirow{3}{*}{HumanEval}
    & Deepseek 1.3B/33B & 4 & 0.70 & 1.05 \\
    & Deepseek 6.7B/33B & 4 & 0.70 & 1.05 \\
    & LLaMA2 7B/70B     & 4 & 0.65 & 1.02 \\
\midrule
\multirow{3}{*}{MBPP}
    & Deepseek 1.3B/33B & 3 & 0.70 & 1.05 \\
    & Deepseek 6.7B/33B & 3 & 0.70 & 1.05 \\
    & LLaMA2 7B/70B     & 3 & 0.70 & 1.05 \\
\midrule
\multirow{3}{*}{CNN/DM}
    & Deepseek 1.3B/33B & 2 & 0.60 & 1.05 \\
    & Deepseek 6.7B/33B & 2 & 0.60 & 1.05 \\
    & LLaMA2 7B/70B     & 3 & 0.65 & 1.05 \\
\midrule
\multirow{3}{*}{GSM8K}
    & Deepseek 1.3B/33B & 4 & 0.65 & 1.05 \\
    & Deepseek 6.7B/33B & 3 & 0.65 & 1.05 \\
    & LLaMA2 7B/70B     & 3 & 0.65 & 1.05 \\
\bottomrule
\end{tabular}
\caption{Block size \(b\), threshold \(t\), and growth factor \(\rho\) used in \method{} across different tasks and models in our experiments.}
\label{tab:eval_params}
\end{table}

\section{Training Details}\label{sec:training_detail}

\begin{figure}[t]
    \centering
    \includegraphics[width=0.5\linewidth]{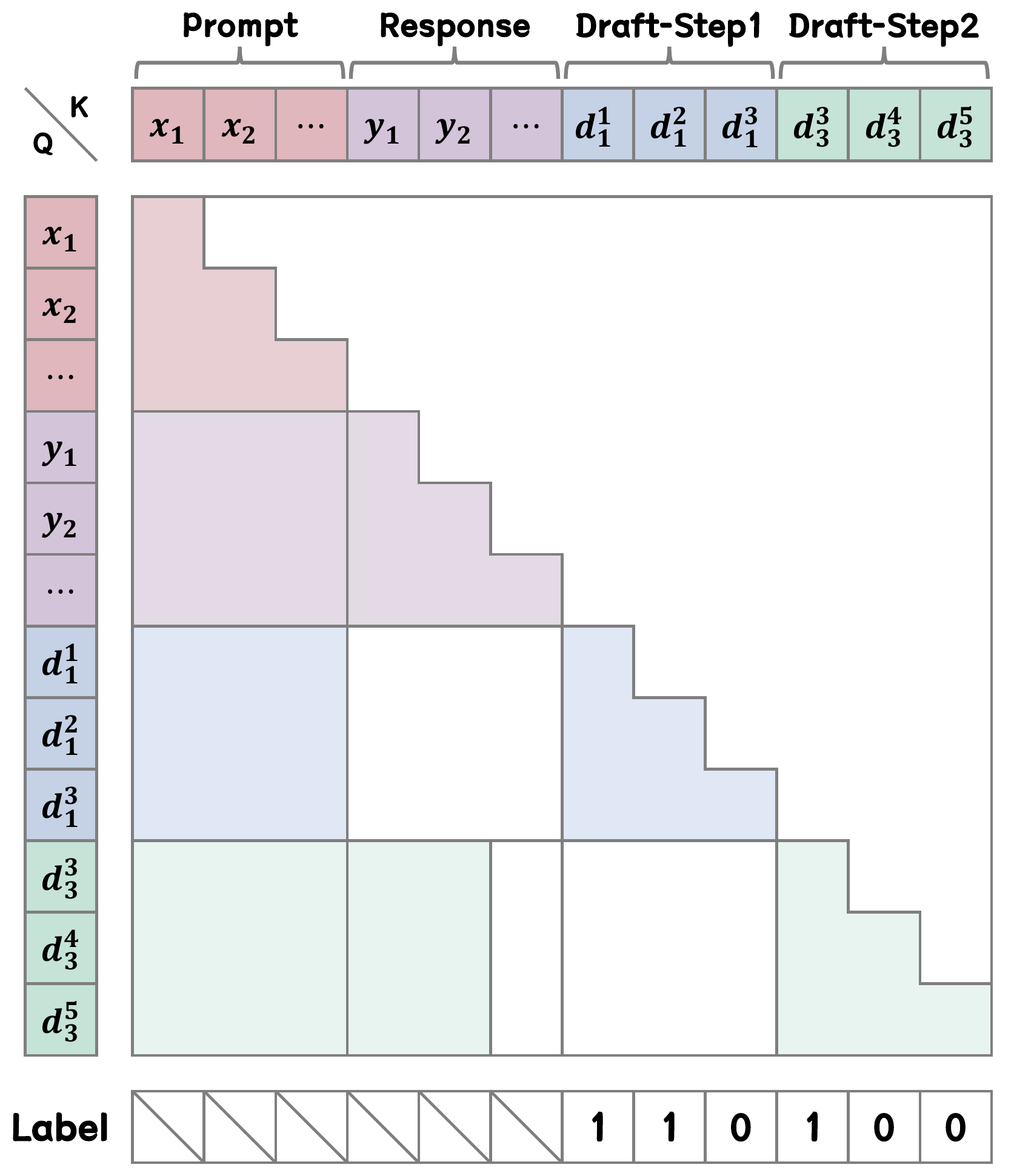}
    \caption{Customized attention mask for packing multiple draft steps into a single sequence for efficient training.}
    \label{fig:attention_mask}
\end{figure}
As discussed in \Cref{sec:training}, we use the same speculative decoding (SD) process—precisely aligned with inference—to construct labeled training data. In this setup, each draft step produces a training example consisting of the input prefix and draft tokens labeled as either accepted or rejected. However, this approach is highly inefficient, as it results in approximately \textcolor{codeblue}{\textbf{\(\text{Len(dataset)} \times \text{Average draft steps}\)}} training examples.

We observe that many draft steps share common prefixes from the same prompt and response, creating redundancy. To address this, we propose packing the training data for all draft steps corresponding to the same prompt into a single training sequence. We apply a customized attention mask during training, as illustrated in \Cref{fig:attention_mask}. Specifically, the draft tokens \(d_i^i, \dots, d_i^{\gamma}\) represent a draft step starting from the prefix \(x_1, x_2, \dots, y_{i-1}\), followed by \(\gamma\) sequentially generated drafts aligned to response positions \(i, \dots, \gamma\). Each draft token \(d_i^m\) is allowed to attend only to the prefix \(x_1, x_2, \dots, y_{i-1}\) and preceding draft tokens \(d_i^i, \dots, d_i^m\).

\section{Additional Ablation Studies}\label{sec:additional_ablations}
We conduct additional ablation studies on block size \(b\), threshold \(t\), and growth factor \(\rho\) using the HumanEval dataset with DeepSeek-Coder 1.3B/33B. We report the decoding speed in tokens per second (tokens/s) for each configuration.

\paragraph{Block Size $b$.} As shown in \cref{tab:block size b}, When the block size $b$ is too small, the accumulation of errors may lead to premature termination of pre-verification. Conversely, when the block size $b$ is too large, the model may fail to trigger early termination via pre-verification, resulting in unnecessary computational overhead due to generating excessive drafts.
\begin{table}[H]
    \small
    \centering
    \begin{tabular}{cccccccc}
        \toprule
        \textbf{$b$} & 1 & 2 & 3 & 4 & 5 & 6 & 7 \\
        \midrule
        \textbf{tokens/s} & 30.00 & 38.07 & 39.28 & 41.80 & 38.95 & 38.85 & 37.21 \\
        \bottomrule
    \end{tabular}
    \caption{Ablations on block size \(b\) where \(b\in[1,7]\)}
    \label{tab:block size b}
\end{table}

\paragraph{Growth Factor $\rho$.} We observe that most training examples involve shorter drafts, making it difficult for the pre-verification layer to generalize well to longer positions. In such cases, the model tends to under-predict the acceptance rate as the position grows. We conduct ablation study to demonstrate the positive effect of growth factors on decoding speed.. As show in \Cref{tab:growth factor}, the heuristic growth factor (e.g., 1.05) serves as a lightweight inductive bias to encourage more aggressive drafting over time, while avoiding unbounded draft lengths.

\begin{table}[H]
    \small
    \centering
    \begin{tabular}{ccccccccccc}
        \toprule
        \textbf{$\rho$} & 1.00 & 1.01 & 1.02 & 1.03 & 1.04 & 1.05 & 1.06 & 1.07 & 1.08 & 1.09 \\
        \midrule
        \textbf{tokens/s} & 40.05 & 40.12 & 40.40 & 40.57 & 40.73 & 41.80 & 41.60 & 40.52 &39.48 & 39.37  \\
        \bottomrule
    \end{tabular}
    \caption{Ablations on growth factor \(\rho\) where \(\rho\in[1.00,1.09]\)}
    \label{tab:growth factor}
\end{table}

\paragraph{Note.} The results for $\rho=1.00$ in~\Cref{tab:growth factor} and for \method{} w/o growth factor in the ablation study of main paper are theoretically equivalent, but differ slightly in implementation. For \method{} w/o growth factor, the threshold remains constant during decoding. For~\Cref{tab:growth factor} ($\rho=1.00$), the threshold is multiplied by $\rho=1.00$ at each step. Although this scaling should have no effect, minor discrepancies arise from subtle control flow or system-level optimizations between the two settings.

\paragraph{Threshold $t$.} We evaluated a series of pre-verfication layers under different threshold $t$ settings, as shown in the \Cref{tab:threshhold t}. A threshold of $0.7$ yields the best overall performance. Setting the threshold too high may lead to premature termination of pre-verification, while a threshold that is too low results in additional overhead from generating unnecessary draft tokens.

\begin{table}[H]
    \small
    \centering
    \begin{tabular}{cccccccc}
        \toprule
        \textbf{$t$} & 0.50 & 0.55 & 0.60 & 0.65 & 0.70 & 0.75 & 0.80 \\
        \midrule
        \textbf{tokens/s} & 37.64 & 38.09 & 40.34 & 39.89 & 41.80 & 40.44 & 39.93 \\
        \bottomrule
    \end{tabular}
    \caption{Ablations on threshold \(t\) where \(t\in[0.50,0.80]\)}
    \label{tab:threshhold t}
\end{table}

\section{Additional Results}
\subsection{Forward Latency}
As shown in \Cref{tab:latency}, the average forward-pass latencies for the blockwise pre-verification layer, draft model, and target model (DeepSeek-Coder 1.3B/33B) are \(1.81\) ms, \(16.52\) ms, and \(67.31\) ms, respectively. The additional overhead introduced by blockwise pre-verification is modest, approximately \(0.11\times\) compared to the draft model and \(0.027\times\) compared to the target model forward pass.

  \begin{table}[h]
    \small
    \centering
    \begin{tabular}{l r}
        \toprule
        \textbf{Model / Layer} & \textbf{Latency (ms)} \\
        \midrule
        Blockwise Pre-verification Layer & 1.81 \\
        \midrule
        DeepSeek-Coder 1.3B & 16.52 \\
        DeepSeek-Coder 33B  & 67.31 \\
        \bottomrule
    \end{tabular}
    \caption{Latency of a single forward pass for different model/layer in DeepSeek-Coder 1.3B/33B using \method{}.}
    \label{tab:latency}
\end{table}

\subsection{Runtime Break of \method{}}
We report the actual runtime breakdown on HumanEval dataset across components for both DeepSeek-Coder 1.3B/33B and Llama-2 7B/70B models, summarized in~\cref{tab:component_analysis}. As shown, the pre-verification layer contributes only \textbf{2.10\%} and \textbf{1.30\%} of the total inference time, respectively. Given the substantial speedups achieved through dynamic draft lengths (72s and 221s), this overhead is moderate in practice.

\begin{table}[h!]
\centering
\small
\begin{tabular}{l c c c c}
\toprule
\multirow{2}{*}{\textbf{Component}} & \multicolumn{2}{c}{\textbf{DeepSeek-Coder 1.3B/33B}} & \multicolumn{2}{c}{\textbf{Llama-2 7B/70B}} \\
\cmidrule(lr){2-3} \cmidrule(lr){4-5}
 & \textbf{Time (s)} & \textbf{Portion (\%)} & \textbf{Time (s)} & \textbf{Portion (\%)} \\
\midrule
Pre-verification Layer & 12.86 & \textbf{2.10} & 22.80 & \textbf{1.30} \\
Draft Model            & 434.55 & 71.24 & 1080.06 & 61.51 \\
Target Model           & 162.59 & 26.66 & 653.14 & 37.19 \\
\midrule
\textbf{Total (\method{})} & \textbf{610} & \textbf{100} & \textbf{1756} & \textbf{100} \\
Total (Speculative)        & 682 & --- & 1977 & --- \\
\bottomrule
\end{tabular}
\caption{Time and portion analysis comparing DeepSeek-Coder and Llama-2 models.}
\label{tab:component_analysis}
\end{table}

\subsection{Detailed Results of Speculative Decoding}\label{sec:sd_results}
\begin{table}[h!]
\centering
\small
\setlength\tabcolsep{3.5pt}
\begin{tabular}{lcccccccccc}
\toprule
\textbf{Model Series} & \textbf{1} & \textbf{2} & \textbf{3} & \textbf{4} & \textbf{5} & \textbf{6} & \textbf{7} & \textbf{8} & \textbf{9} & \textbf{10} \\
\midrule
DeepSeek-Coder 1.3B/33B & -- & -- & -- & 34.99 & 36.11 & 37.27 & 37.41 & 36.69 & \textbf{37.46} & 37.11 \\
DeepSeek-Coder 6.7B/33B & 26.83 & 28.81 & 29.35 & 29.57 & \textbf{29.77} & 29.31 & 29.70 & 29.49 & 29.41 & -- \\
Qwen2.5 1.5B/32B        & 24.07 & 27.29 & 28.14 & \textbf{28.46} & 27.78 & 27.77 & 27.35 & 26.97 & 26.32 & -- \\
LLama2 7B/70B      & -- & -- & -- & 18.59 & 20.95 & 21.15 & 21.21 & \textbf{21.23} & 21.10 & 21.12 \\
\bottomrule
\end{tabular}
\caption{Decoding speeds (tokens/s) for SD of various fixed window sizes (\(\gamma = 1\) to \(\gamma = 10\)) across different model series.}
\label{tab:draft_length}
\end{table}
For vanilla speculative decoding, we test different numbers of draft tokens as shown in \Cref{tab:draft_length}, and we choose the optimal result as a baseline for our method.

\subsection{Acceptance Rates Across Draft Positions}
\begin{figure}[h!]
    \centering
    \includegraphics[width=0.8\linewidth]{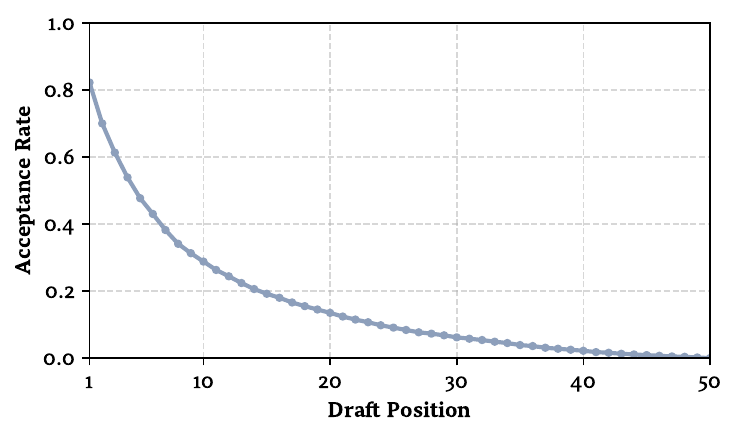}
    \caption{Acceptance rates with different draft positions.}
    \label{fig:draft_pos}
\end{figure}

\section{Limitations}

Although the proposed \method{} effectively controls draft lengths dynamically and accelerates speculative decoding, it still has certain limitations. Its performance depends on the quality of drafts produced by the draft model. For tasks like code generation (e.g., HumanEval), longer average acceptance lengths yield substantial speedups; however, for summarization tasks (e.g., CNN/DM), optimal drafts tend to be shorter, limiting potential performance gains. Similarly, when both draft and target models are relatively small, the optimal draft length is short, and the relative overhead of the pre-verification layer may become more pronounced. Nevertheless, \method{} complements speculative decoding techniques aimed at enhancing draft quality. Integrating \method{} with Ouroboros~\citep{zhao2024ouroboros} further boosts decoding speed by up to $3.09\times$ over autoregressive decoding. We anticipate that future research in speculative decoding will continually enhance draft quality. With such advancements, our approach is expected to achieve even more significant performance improvements.

\end{document}